\documentclass[10pt,twocolumn,letterpaper]{article}

\usepackage{graphicx}
\usepackage{amsmath}
\usepackage{amssymb}
\usepackage{booktabs}

\usepackage{epsfig}
\usepackage{graphicx}
\usepackage{amsmath}
\usepackage{amssymb}
\usepackage{multirow}
\usepackage{enumerate}
\usepackage{bm}
\usepackage{wrapfig}
\usepackage{epstopdf}
\usepackage{wrapfig}

\usepackage{tabularx}
\usepackage{graphicx}
\usepackage{adjustbox}
\usepackage{makecell}

\DeclareMathOperator{\argmin}{argmin} 

\usepackage[pagebackref,breaklinks,colorlinks]{hyperref}

\usepackage[capitalize]{cleveref}
\crefname{section}{Sec.}{Secs.}
\Crefname{section}{Section}{Sections}
\Crefname{table}{Table}{Tables}
\crefname{table}{Tab.}{Tabs.}

\usepackage{iccv}
\usepackage{times}
\usepackage{epsfig}
\usepackage{graphicx}
\usepackage{amsmath}
\usepackage{amssymb}
\usepackage{tabularx}
\usepackage{makecell} 
\usepackage{epsfig}
\usepackage{graphicx}
\usepackage{amsmath}
\usepackage{amssymb}
\usepackage{multirow}
\usepackage{enumerate}
\usepackage{bm}
\usepackage{wrapfig}
\usepackage{epstopdf}
\usepackage{wrapfig}

 \iccvfinalcopy 

\def\iccvPaperID{3424} 

\ificcvfinal\fi

\begin{document}

\title{Coordinate Quantized Neural Implicit Representations for Multi-view Reconstruction}


\author{Sijia Jiang, Jing Hua, Zhizhong Han\\
Department of Computer Science, Wayne State University, Detroit, USA\\
{\tt\small sijiajiang@wayne.edu, jinghua@wayne.edu, h312h@wayne.edu}
}
\maketitle
\ificcvfinal\thispagestyle{empty}\fi
\begin{abstract}
In recent years, huge progress has been made on learning neural implicit representations from multi-view images for 3D reconstruction. As an additional input complementing coordinates, using sinusoidal functions as positional encodings plays a key role in revealing high frequency details with coordinate-based neural networks. However, high frequency positional encodings make the optimization unstable, which results in noisy reconstructions and artifacts in empty space. To resolve this issue in a general sense, we introduce to learn neural implicit representations with quantized coordinates, which reduces the uncertainty and ambiguity in the field during optimization. Instead of continuous coordinates, we discretize continuous coordinates into discrete coordinates using nearest interpolation among quantized coordinates which are obtained by discretizing the field in an extremely high resolution. We use discrete coordinates and their positional encodings to learn implicit functions through volume rendering. This significantly reduces the variations in the sample space, and triggers more multi-view consistency constraints on intersections of rays from different views, which enables to infer implicit function in a more effective way. Our quantized coordinates do not bring any computational burden, and can seamlessly work upon the latest methods. Our evaluations under the widely used benchmarks show our superiority over the state-of-the-art. Our code is available at \href{
https://github.com/MachinePerceptionLab/CQ-NIR}{https://github.com/MachinePerceptionLab/CQ-NIR}.
\end{abstract}

\section{Introduction}
Learning implicit representations from multi-view images is a challenge in reconstructing 3D geometry in a scene. The latest methods learn implicit representations using coordinate-based neural networks to infer signed distance or occupancy fields~\cite{Oechsle2021ICCV,yariv2020multiview,yariv2021volume,geoneusfu,neuslingjie,Yu2022MonoSDF,yiqunhfSDF,Vicini2022sdf,wang2022neuris,guo2022manhattan,li2023neuralangelo,rosu2023permutosdf} through volume rendering. By shooting rays across the fields, we render RGB values at a pixel by integrating colors and geometry at 3D queries sampled along a ray through volume rendering. The images rendered from neural implicit functions are compared with the ground truth images, which measures errors to improve the neural implicit fucntions.

Learning high fidelity implicit representations requires to use positional encodings~\cite{mildenhall2020nerf,Oechsle2021ICCV,neuslingjie,yiqunhfSDF} as a complement to coordinates, which remedies the incapability of coordinate-based neural networks in modeling high frequency details. Positional encodings are vectors formed by sinusoidal functions of coordinates with both low and high frequencies~\cite{sitzmann2019siren,tancik2020fourfeat}, where the frequency band is shown as the key factor to capture details in different scenes. However, higher frequency turns out to bring noises, which results in artifacts on surfaces and in empty spaces. To stabilize the optimization with high frequency positional encodings, some methods~\cite{hertz2021sape,yiqunhfSDF,park2021nerfies} learn soft masks to gradually expose high frequency components over training iterations. However, this masking strategy relies on training iterations and numbers of frequency components, which is tedious to tune in a general sense.

To resolve this issue, we propose to use quantized coordinates to learn neural implicit representations from multi-view images. Instead of continuous coordinates and positional encodings of continuous coordinates in previous methods~\cite{mildenhall2020nerf,neuslingjie,Yu2022MonoSDF,yiqunhfSDF,Vicini2022sdf,wang2022neuris,guo2022manhattan,jun2023learning}, we use discrete coordinates and positional encodings of discrete coordinates as the input of coordinate-based neural networks, where we discretize the field in an extremely high resolutions. Our insight here is to decrease the uncertainty and ambiguity in the field during optimization. We achieve this by introducing discrete coordinates with two reasons. On the one hand, we enable networks to merely observe a finite set of discrete coordinates rather than infinite continuous variations, which simplifies the optimization by significantly reducing variations in the sample space. On the other hand, a discrete coordinate covers an area rather than a point, hence rays from different views are more easily to have overlapped samples with each other. This triggers more multi-view consistency constraints to take effect at these intersections, which leads to more effective inference. Our quantized coordinates do not bring any extra computational burden, inconsistency on borders of neighboring coordinates, and provide a general strategy which can be used upon different methods. We evaluate our improvements over the latest methods under multiple benchmarks. Our contributions are listed below.\vspace{-0.1in}

\begin{enumerate}[i)]
\item We introduce quantized coordinates to learn neural implicit functions from multi-view images. By discretizing a field in an extremely high resolution, we introduce efficient ways of using discrete coordinates, which does not bring extra computational burden and inconsistency on borders of neighboring coordinates.\vspace{-0.1in}
\item We report analysis on how discrete coordinates decrease the uncertainty and ambiguity in the field by reducing the variations in the sample space and triggering more multi-view consistency constraints to infer implicit functions in a more effective way.\vspace{-0.1in}
\item Our discrete coordinates can seamlessly work upon the latest methods. We justify our effectiveness by showing significant improvements over the state-of-the-art results under the widely used benchmarks.
\end{enumerate}

\section{Related Work}

\noindent\textbf{3D Reconstruction from Multiple Images. }Reconstructing 3D shapes from multiple images has been extensively studied in 3D computer vision~\cite{schoenberger2016sfm,schoenberger2016mvs,mildenhall2020nerf,GEOnEUS2022,Oechsle2021ICCV,neuslingjie,Yu2022MonoSDF,yiqunhfSDF,Vicini2022sdf,wang2022neuris,guo2022manhattan,chen2023fantasia3d,reiser2023merf}. Given multiple RGB images, classic multi-view stereo (MVS)~\cite{schoenberger2016sfm,schoenberger2016mvs} methods employ multi-view consistency to estimate depth information. They rely on matching key points on different views, which is limited by large viewpoint variations and complex illumination. With multiple silhouette images, we can reconstruct 3D shapes as voxel grids using space carving~\cite{273735visualhull}. The disadvantages of these methods include the inability of revealing concave structures and low resolutions in voxel grids.

Recent methods~\cite{yao2018mvsnet} employ neural networks to implement the MVS framework. During training, they learn priors using depth supervision or multi-view consistency in an unsupervised way, and then, generalize the priors to predict depth images for unseen cases through a forward pass.

These methods reconstructed 3D shapes as point clouds or voxel grids, both of which are discrete. While neural implicit representations for 3D reconstruction represent surfaces as the level set which is continuous.

\noindent\textbf{Neural Implicit Representations. }Neural implicit representations have shown prominent performance in representing 3D geometry~\cite{DBLP:journals/corr/abs-1901-06802,Park_2019_CVPR,MeschederNetworks,chen2018implicit_decoder,Jiang2019SDFDiffDRcvpr,chaompi2022,jun2023shape,poole2022dreamfusion}. We can learn neural implicit representations using coordinate-based neural networks from 3D supervision~\cite{jiang2020lig,DBLP:conf/eccv/ChabraLISSLN20,Peng2020ECCV,DBLP:journals/corr/abs-2105-02788,takikawa2021nglod,Liu2021MLS,tang2021sign}, point clouds~\cite{Zhou2022CAP-UDF,Zhizhong2021icml,DBLP:conf/icml/GroppYHAL20,Atzmon_2020_CVPR,zhao2020signagnostic,atzmon2020sald,chaompi2022,nichol2022pointe,gupta20233dgen,chao2023grid,ma2023learning,chen2023unsupervised,Zhou2022CAP-UDF,melaskyriazi2023pc2,predictivecontextpriors2022,onsurfacepriors2022,DBLP:conf/cvpr/LiWLSH22}, or multi-view images~\cite{mildenhall2020nerf,GEOnEUS2022,Oechsle2021ICCV,neuslingjie,Yu2022MonoSDF,yiqunhfSDF,Vicini2022sdf,wang2022neuris,guo2022manhattan,meng2023neat}. Since 3D supervision and point clouds expose more explicit geometry clues than multiple images, methods learning from these kinds of supervision do not employ positional encodings as input. Hence, our discrete coordinates are mainly evaluated upon the methods using multi-view images as supervision.

With differentiable rendering techniques, we are enabled to evaluate the correctness of neural implicit representations using errors between rendered images and ground truth images. With surface rendering~\cite{Jiang2019SDFDiffDRcvpr}, DVR~\cite{DVRcvpr} and IDR~\cite{yariv2020multiview} infer the radiance on surfaces. IDR also models view direction as a condition to reconstruct high frequency details. Since these methods focus on surfaces, they require masks to filter out the background.

NeRF~\cite{mildenhall2020nerf} and its variations~\cite{park2021nerfies,mueller2022instant,ruckert2021adop,yu_and_fridovichkeil2021plenoxels,raj2023dreambooth3d,liu2023neudf,melaskyriazi2023realfusion} use volume rendering to simultaneously model geometry and color. These methods were proposed for novel view synthesis, and render images without masks. Using volume rendering, unisurf~\cite{Oechsle2021ICCV} and NeuS~\cite{neuslingjie} revise the rendering procedure to render occupancy and signed distance fields with colors, which infers accurate implicit functions. Following methods improve accuracy of implicit functions using additional priors or losses including depth~\cite{Yu2022MonoSDF,Azinovic_2022_CVPR,Zhu2022CVPR}, normals~\cite{Yu2022MonoSDF,wang2022neuris,guo2022manhattan}, and multi-view consistency~\cite{GEOnEUS2022}.

\section{Preliminary}
\noindent\textbf{Neural Radiance Fields. }NeRF~\cite{mildenhall2020nerf} represents scenes by jointly modeling volume densities and colors using a neural network. Starting a pixel, we shoot a ray, and integrates densities and colors at samples along the ray into RGB values at the pixel through volume rendering. At a 3D sample $\bm{q}\in\mathbb{R}^3$, the neural network predicts the density $\sigma(\bm{q})\in\mathbb{R}^+$ and color $c(\bm{q},\bm{d})\in\mathbb{R}^3$, where $\bm{d}$ indicates the ray direction passing $\bm{q}$ which enables to model view-dependent effects such as reflections.

To sample $I$ queries $\{\bm{q}_i\}$ along a ray with a direction $\bm{d}$, NeRF parameterizes the ray using the distance $t$ to the camera center $\bm{o}$, $\bm{q}_i=\bm{o}+t\bm{d}$. The rendered color along the ray is obtained by volume rendering below,\vspace{-0.05in}

\begin{equation}
\label{eq:rendering1}
C=\sum_{i=1}^IT_i(1-exp(-\sigma(\bm{q}_i)\delta_i))c(\bm{q}_i,\bm{d}),
\vspace{-0.1in}
\end{equation}

\noindent where $T_i=exp(-\sum_{j<i}\sigma(\bm{q}_j)\delta_j)$ is the accumulated transmittance along the ray and $\delta_i$ is the Euclidean distance between $\bm{q}_{j+1}$ and $\bm{q}_{j}$. The network can be trained by minimizing the error between rendered images and ground truth images through the differentiable volume rendering.

To better fit data containing high frequency variations, positional encoding is introduced to map coordinates $\bm{q}_i$ into a higher dimensional space using sinusoidal functions with a frequency band. Formally, the encoding function $\gamma(\bm{q}_i)$ is defined below,\vspace{-0.1in}

\begin{equation}
\label{eq:encoding}
(sin(\omega_1\bm{q}_i),cos(\omega_1\bm{q}_i),...,sin(\omega_{L}\bm{q}_i),cos(\omega_{L}\bm{q}_i)),
\vspace{0.1in}
\end{equation}

\noindent where $\{\omega_1,...,\omega_{L}\}$ is a band containing $L$ frequencies, and $\omega_{L}=2^{L-1}\pi$. The encoding function $\gamma$ is applied to each element in the coordinate vector $\bm{q}_i$ and the vector indicating view direction $\bm{d}$.

\noindent\textbf{Neural Implicit Representations. }Based on NeRF, the latest methods learn neural implicit representations $f_{\theta}(\bm{q}_i)$, such as occupancy fields~\cite{MeschederNetworks} and signed distance fields~\cite{Park_2019_CVPR}, along with a color value $c_{\theta}(\bm{q}_i,\bm{d})$, through volume rendering. These methods reformulate the volume rendering equation in Eq.~\ref{eq:rendering1} to replace densities into occupancy labels or signed distances, where a function $V$ is defined to map $f_{\theta}$ into alpha. The function $V$ is the key to enable to learn 3D implicit representations from 2D supervision without masks. These methods optimize neural networks parameterized by $\theta$ by minimizing the squared error below,\vspace{-0.1in}

 \begin{equation}
\label{eq:error}
\min_{\theta} ||C_{GT}-\sum_i^IV(\{f_{\theta}(\bm{q}_{j\le i})\})C_{\theta}(\bm{q}_i,\bm{d})||_2^2,
\end{equation}

\noindent where $C_{GT}$ is the ground truth color at the pixel emitting the ray.

\begin{figure}[tb]
  \centering
   \includegraphics[width=\linewidth]{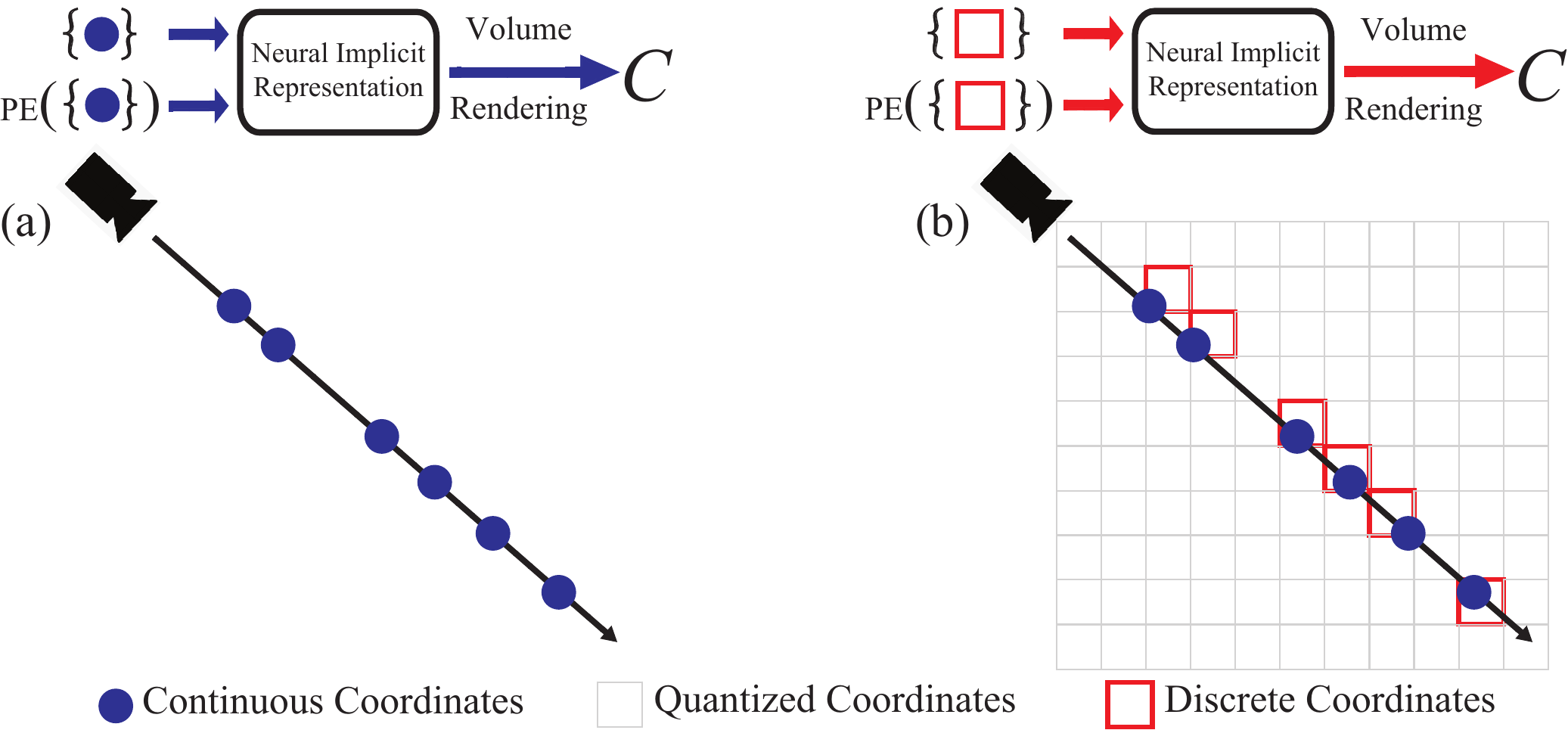}
  %
  %
\vspace{-0.1in}
\caption{\label{fig:overview}Illustration of our method. We discretize continuous coordinates in (a) into discrete coordinates using quantized coordinates in an extremely high resolution in (b). We use these discrete coordinates and their positional encoding to learn neural implicit representations. }
\vspace{-0.2in}
\end{figure}

Moreover, these methods improve the sampling strategy to infer implicit representations in a more efficient way. Specifically, they first use ray marching or secant method to find the intersection between the ray and the scene, and then sample more queries around the intersection to do volume rendering, where intersections are estimated surface points. More advanced techniques for smoother implicit fields include using normals of surface points as the input of the color network~\cite{Oechsle2021ICCV,yariv2020multiview,yariv2021volume,geoneusfu,neuslingjie}, adding constraints on normals of neighboring points~\cite{Oechsle2021ICCV}, and using sparse depth from MVS as priors~\cite{GEOnEUS2022}.

\section{Method}
\noindent\textbf{Issues of Continuous Coordinates. }To render RGB values along a ray, current methods sample queries $\bm{q}_i$ along the ray according to some specific sampling strategy. As illustrated in Fig.~\ref{fig:overview} (a), these queries are used as probes to sense the continuous field for volume rendering. They are associated with continuous coordinates which are further manipulated by sinusoidal functions with a frequency band as a positional encoding $\gamma(\bm{q}_i)$. Continuous coordinates have two issues.

\begin{figure}[tb]
  \centering
\includegraphics[width=0.5\linewidth]{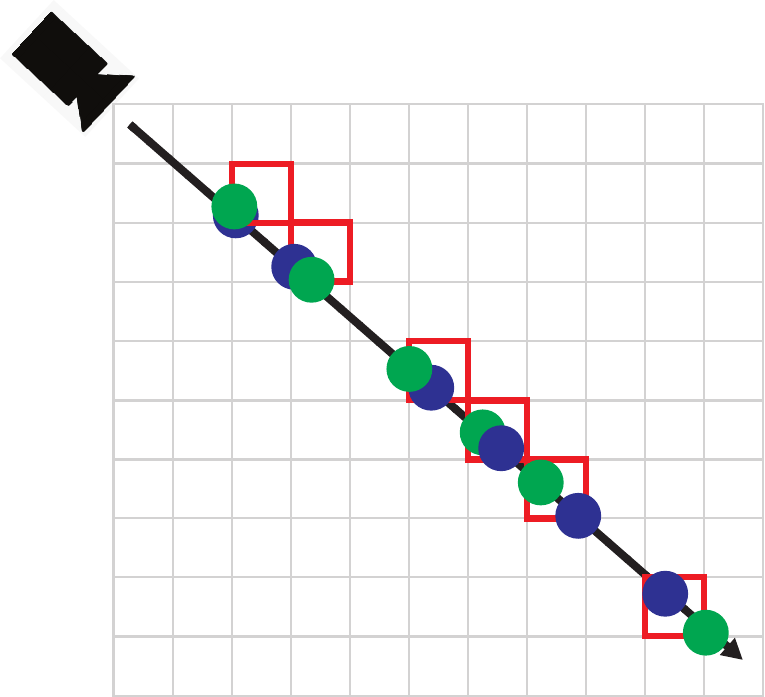}
\caption{\label{fig:Variations}Different colors indicate different iterations (shown as nodes in two colors) of the same ray. Along the same ray, continuous coordinates sampled in different iterations are different, while their discrete coordinates may be the same set.  }
\end{figure}

For one thing, continuous coordinates produce a huge variations in the sample space. Along the same ray shown in Fig.~\ref{fig:Variations}, the queries sampled in one iteration (blue dots) do not overlap with the ones sampled in another iterations (green dots) since the field is optimized through instantly tuning parameters of the network. The variations are extended to be even larger with high dimensional positional encodings $\gamma$ as an additional input. This makes implicit functions keep observing different samples as input during training, which is an obstacle that make neural networks to struggle to infer uncertainties and ambiguities in the field.


\begin{figure*}[tb]
  \centering
   \includegraphics[width=\linewidth]{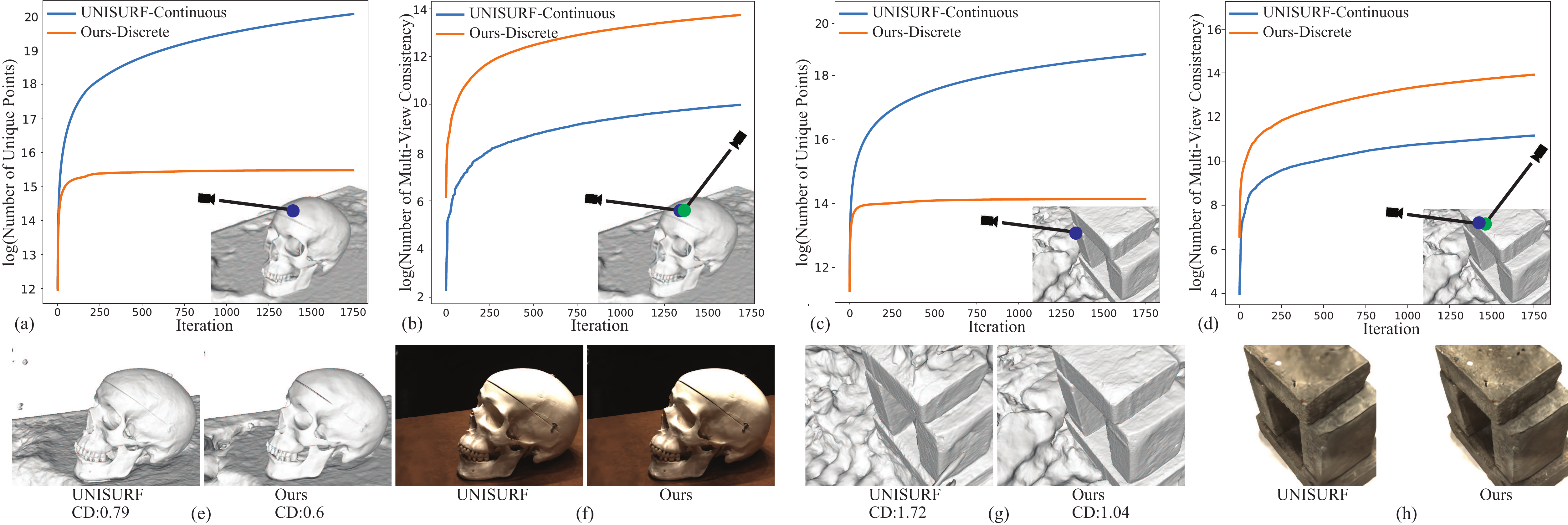}
  %
  %
 \vspace{-0.05in}
\caption{\label{fig:counts}Statistics comparison of continuous coordinates and discrete coordinates in terms of the the number of unique coordinates that the network has observed and the number of overlapped samples along rays from different views.}
\vspace{0.05in}
\end{figure*}

\begin{figure}[tb]
  \centering
   \includegraphics[width=\linewidth]{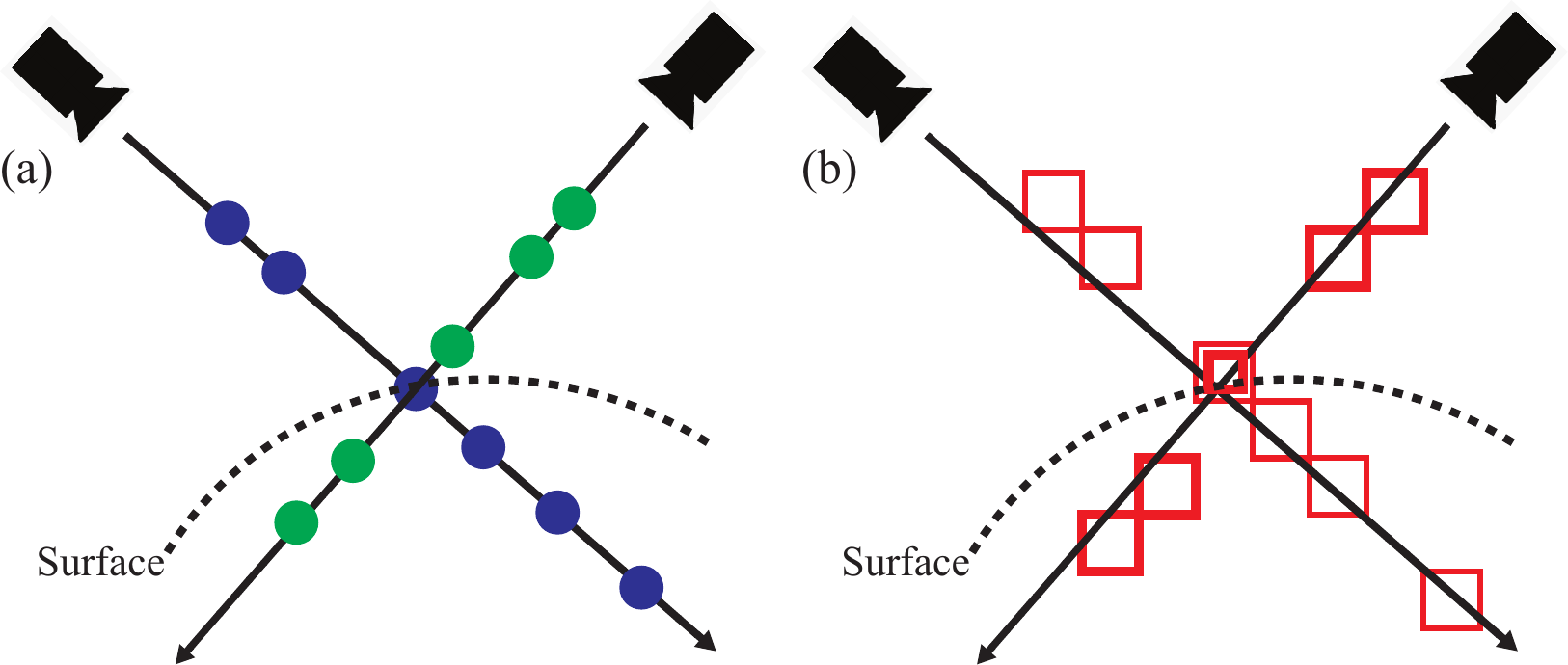}
  %
  %
 \vspace{0.15in}
\caption{\label{fig:MVC}Discrete coordinates in (b) make it easier to involve the same location in volume rendering along the rays from different views than continuous coordinates in (a).}
\vspace{-0.0in}
\end{figure}

For another, continuous coordinates are not effective to impose multi-view consistency constraints on inferring implicit functions. The essence of using multi-view consistency to infer occupancy or signed distances is to involve intersections of rays from different views in volume rendering. However, queries are sampled on rays from different views separately without considering consistency, which may make points sampled on both rays do not overlap at the intersection duo to randomness in sampling. As illustrated in Fig.~\ref{fig:MVC} (a), two rays from two differen views are supposed to intersect on a surface point, while the points sampled for volume rendering along the two rays (blue dots in one ray, green dots in another ray) do not overlap at the intersection. This means that the intersection will not get involved in volume rendering along both rays, resulting in no multi-view consistency constraints to be imposed on the intersection for inferring implicit function value. This is also another obstacle for inferring uncertainties and ambiguities in the field.

\noindent\textbf{Quantized Coordinates. }To resolve the issues of continuous coordinates, we introduce quantized coordinates to learn implicit functions from multi-view images. As shown in Fig.~\ref{fig:overview} (b), we first obtain quantized coordinates (centers of squares) by discretizing the field in an extremely high resolution, and then leverage these quantized coordinates to discretize continuous coordinates (blue dots) into discrete ones (centers of red squares). Specifically, we voxelize the field into a voxel grid with a resolution of $R$, and use the center of each voxel as quantized coordinates $\bm{e}_j\in\mathbb{R}^3$, where $j\in[1,J]$ and $J=R^3$. We denote the set of quantized coordinates as $E=\{\bm{e}_j|j\in[1,R^3]\}$, and use nearest interpolation in $E$ to discretize each query $\bm{q}_i$ into a discrete coordinate $\tilde{\bm{q}_i}$, as formulated by,\vspace{-0.0in}

\begin{equation}
\label{eq:nearest}
\tilde{\bm{q}_i}=\bm{e}_k,\quad \text{where} \quad k=\argmin_j||\bm{q}_i-\bm{e}_j||_2.
\vspace{-0.0in}
\end{equation}

We use discrete coordinates $\tilde{\bm{q}_i}$ of queries and their corresponding positional encodings $\gamma(\tilde{\bm{q}_i})$ as the input of the network. We reformulate Eq.~\ref{eq:encoding} to obtain $\gamma(\tilde{\bm{q}_i})$ as,\vspace{-0.0in}

\begin{equation}
\label{eq:encodingdiscrete}
(sin(\omega_1\tilde{\bm{q}_i}),cos(\omega_1\tilde{\bm{q}_i}),...,sin(\omega_{L}\tilde{\bm{q}_i}),cos(\omega_{L}\tilde{\bm{q}_i}),
\vspace{0.05in}
\end{equation}

\noindent where $\{\omega_1,...,\omega_{L}\}$ is also a band containing $L$ frequencies, and $\omega_{L}=2^{L-1}\pi$. Accordingly, with discrete coordinates $\tilde{\bm{q}_i}$, we reformulate volume rendering as,\vspace{-0.0in}

\begin{equation}
\label{eq:rendering1discrete}
C=\sum_{i=1}^IT_i(1-exp(-\sigma(\tilde{\bm{q}_i})\delta_i))c(\tilde{\bm{q}_i},\bm{d}),
\vspace{-0.0in}
\end{equation}

\noindent where $T_i=exp(-\sum_{j<i}\sigma(\tilde{\bm{q}_i})\delta_j)$ is the accumulated transmittance along the ray and $\delta_i$ is still the Euclidean distance between continuous coordinates $\bm{q}_{j+1}$ and $\bm{q}_{j}$. In addition, we remove the duplicated discrete coordinates along a ray, and use the unique discrete coordinates in volume rendering in Eq.~\ref{eq:rendering1discrete}, which achieves more accurate approximation. We learn neural implicit functions with discrete coordinates by minimizing the rendering error,\vspace{-0.05in}

 \begin{equation}
\label{eq:error1}
\min_{\theta} ||C_{GT}-\sum_i^IV(\{f_{\theta}(\tilde{\bm{q}}_{j\le i})\})C_{\theta}(\tilde{\bm{q}_i},\bm{d})||_2^2.
\vspace{-0.00in}
\end{equation}

\noindent\textbf{Why Discrete Coordinates Work. }With discrete coordinates, we can significantly reduce the variations in the sample space. Obviously, the network observes a finite set of the coordinates and also their corresponding positional encodings, rather than infinite variations with continuous coordinates especially with high frequencies positional encodings. As illustrated in Fig.~\ref{fig:Variations}, for points sampled in different iterations (blue and green dots), their discrete coordinates are the same (indicated by the centers of the red boxes).

We statistically justify our claim in reconstructing a 3D scene from multi-view images in Fig.~\ref{fig:counts} (a). We use UNISURF~\cite{Oechsle2021ICCV} as a baseline which samples continuous coordinates to learn an occupance field via volume rendering. We discrete these continuous coordinates into discrete coordinates using nearest interpolation over $R^3$ quantized coordinates, where $R=51200$. During optimization, we monitor a fixed set of $1024$ rays in each iteration, we focus on the rays that hit the surface, and record the continuous coordinates and their discrete coordinates sampled on these rays. We count the number of unique continuous coordinates and the number of unique discrete coordinates that the network has observed respectively. We accumulate these numbers over training iterations respectively, and show them using a logarithmic function as the two lines in Fig.~\ref{fig:counts} (a). The comparison shows that discrete coordinates overlap a lot in different iterations, so the number of unique discrete coordinates increases very slowly, which expands much smaller variations in the sample space than continuous coordinates. We repeat this experiment in another scene, and we observe the similar statistics in Fig.~\ref{fig:counts} (c).

Moreover, the queries sampled on rays from different views have higher probability of co-occurrence in the same voxel, which makes them share the same discrete coordinate. As illustrated in Fig.~\ref{fig:MVC} (a), although continuous coordinates of samples along two rays (blue and green dots) do not have an overlap at the intersection of the two rays on the surface, their corresponding discrete coordinates (centers of red boxes and centers of red boxes in bold) share the same intersection in Fig.~\ref{fig:MVC} (b). This leads to involve the shared discrete coordinate in the volume rendering along both of the rays, which imposes multi-view consistency constraints at the shared discrete coordinate in a more effective way. More importantly, discrete coordinates facilitate rays from different views more easily intersect at discrete coordinates, which triggers more multi-view consistency constraints.

We statistically justify our claim using the same setting as Fig.~\ref{fig:counts} (a), and count the number of two rays that have an overlapped sampling on the surface in Fig.~\ref{fig:counts} (b). We still use $51200^3$ quantized coordinates to discretize continuous coordinates of samples on rays, and monitor a fixed set of $1024$ rays during optimization. For each ray that hits the surface, we project the hitting point to a neighboring view, and use the projection trajectory as another ray. Then, we sample points along these two rays separately using the sampling strategy in UNISURF, and check whether the two sets of sampled points have an overlap at the intersection. If both of the two rays have a sample on the surface, and their distance is smaller than a small threshold ($1/16$ of a voxel size), we regard these two rays involve the same sampled point in volume rendering, which indicates that a multi-view constraint take effect one time. Similarly, we check whether the two sampled points on the surface have the same discrete coordinates. We accumulate the times of multi-view constraint taking effect over iterations with continuous coordinates or discrete coordinates respectively, and show them using a logarithmic function as the two lines in Fig.~\ref{fig:counts} (b). The comparison shows that discrete coordinates triggers much more multi-view constraints than continuous coordinates. We repeat this experiment in another scene, and we observe the similar statistics in Fig.~\ref{fig:counts} (d). Although neural network can generalize around continuous coordinates, imposing multi-view constraints on the same location through volume rendering can effectively infer 3D geometry with higher accuracy.

These benefits from discrete coordinates are vital to stabilize the optimization by reducing the uncertainty and ambiguity in the field, which achieves to reveal more accurate geometry and more smoother surfaces as the visual comparison in Fig.~\ref{fig:counts} (e) and (g). Although we dicretize the field, the extremely high resolution does not produce artifacts or sawthooth effect in geometry or rendered images in Fig.~\ref{fig:counts} (f) and (h).

\noindent\textbf{Border Consistency. }Our quantized coordinates do not bring the disadvantages of voxelizing a field to neural implicit functions, although our voxelize the field in an extremely high resolution.

The disadvantages of voxelization include cubic computational complexity and inconsistency on borders of neighboring quantized coordinates. Different from feature grids~\cite{DBLP:conf/eccv/PengNMP020,Zhu2022CVPR}, which hold learnable features at vertices of grids in memory, we calculate quantized coordinates using a function of the field range and resolution, which does not bring any storage burden. This is also the key to enable us to quantize coordinates in an extremely high resolution. In addition, we implement the nearest interpolation in Eq.~\ref{eq:nearest} by getting a continuous coordinate divided by the voxel interval, which avoids the computational nearest search. To achieve border consistency, the methods of learning local implicit functions~\cite{jiang2020lig,DBLP:conf/eccv/ChabraLISSLN20} use trilinear interpolation to interpolate features~\cite{DBLP:conf/eccv/PengNMP020,Zhu2022CVPR} or implicit function values~\cite{jiang2020lig} from the nearest 8 voxel vertices. In contrast, our extremely high resolution leads to very small interval between neighboring quantized coordinates, which almost brings no inconsistency in implicit functions values or degenerate the rendering, as illustrated in Fig.~\ref{fig:counts} (e) and (g).

\noindent\textbf{Resolutions. }Although we claim quantized coordinates in an extremely high resolution benefit the learning of neural implicit representations from multi-view images, we note that continuous coordinates are actually quantized coordinates in an infinite resolution. Hence, a too high resolution does not help improve the inference. We will explore the effect of resolutions in experiments.

\section{Experiments}
We evaluate our method in 3D reconstruction from multi-view images for shapes with background and large scale scenes. We use quantized coordinate to learn either signed distance fields or occupancy fields with different baselines, and then run the marching cubes algorithm~\cite{Lorensen87marchingcubes} to extract the zero level set as a surface. Note that we also use discrete coordinates to produce discrete signed distance field for the marching cubes.

\subsection{Evaluations for Shapes}
\noindent\textbf{Dataset and Metrics. }We evaluate our method in reconstructing 3D shapes without masks using multi-view images from the DTU datatset~\cite{jensen2014large}. Following previous methods~\cite{Oechsle2021ICCV,yariv2020multiview,yariv2021volume,geoneusfu,neuslingjie,Yu2022MonoSDF,yiqunhfSDF}, we report our performance on the widely used $15$ scans. For each scan, a scene is represented by $49$ to $64$ images with different shape appearances.

We use Chamfer distance to evaluate the accuracy of reconstructed surfaces, where we randomly sample points on the reconstructed surfaces, and compare them to the ground truth. Following previous methods~\cite{Oechsle2021ICCV,yariv2020multiview,yariv2021volume,geoneusfu,neuslingjie,Yu2022MonoSDF,yiqunhfSDF,chou2022gensdf}, we clean the reconstructed meshes using the respective masks. We use the official evaluation code released by the DTU dataset to measure our accuracy.

\noindent\textbf{Baselines. }To evaluate our method in learning both singed distance field and occupancy field, we use UNISURF~\cite{Oechsle2021ICCV}, NeuS~\cite{neuslingjie}, Geo-Neus~\cite{GEOnEUS2022} and NeuralWarp~\cite{DBLP:conf/cvpr/DarmonBDMA22} as baselines which are the state-of-the-art methods for learning implicit functions from multi-view images. All these methods do not use priors. Moreover, we do not evaluate our method in novel view synthesis, since the shape and radiance ambiguity~\cite{Zhang20arxiv_nerf++} makes geometry no need to be represented as a surface, which is hard to have overlapped samples along different rays. 


\noindent\textbf{Details. }We discretize a field into $R^3$ voxels in an extremely high resolution, and regard the center of each voxel as a quantized coordinate. For the range of a field, UNISURF, NeuS, and NeuralWarp normalize a scene into a cube with a range of $[-4,4]$, $[-2.5,2.5]$, and $[-5.5,5.5]$, respectively. To evaluate our methods with different resolutions, we evaluate our results with two resolution settings which keeps each quantized coordinate covering a area with a similar size, i.e., $R=51200$ and $R=25600$ for UNISURF and NeuS, $R=70400$ and $R=51200$ for NeuralWarp. We report these results in our supplementary materials, and list summarized results in the main text.

We use the official code released by UNISURF, NeuS, and NeuralWarp to produce our results with discrete coordinates. Moreover, we use the corresponding discrete coordinates to calculate positional encodings as in Eq.~\ref{eq:encodingdiscrete}. For the normals required for color prediction or loss calculation in these methods, we also use the normals at the discrete coordinates. For the warping in NeuralWarp, we still use the continuous coordinates to get precise color from other views.

\begin{figure*}[tb]
  \centering
   \includegraphics[width=\linewidth]{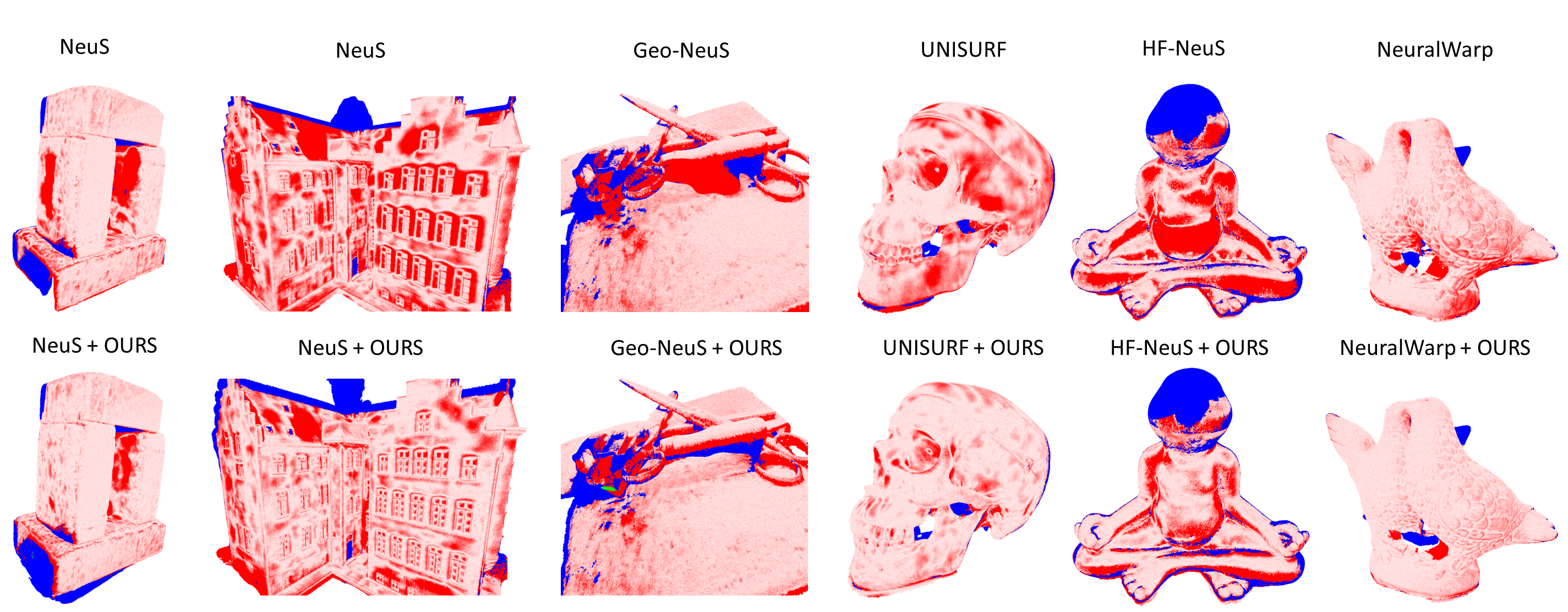}
  %
  %
 \vspace{-0.0in}
\caption{\label{fig:error_maps}Visual comparison on DTU. Error maps 
highlight our improvements (white to red) over different baselines.}
\vspace{-0.0in}
\end{figure*}

\noindent\textbf{Comparison. }We report numerical evaluations in DTU in Table.~\ref{table:dtu}. We improve the performance of our baselines including UNISURF, NeuS, and NeuralWarp. Specifically, our results in all scenes outperform UNISURF. Except our comparable result in scene 97, we also achieve better performance than NeuralWarp in other scenes. Using NeuS as a baseline, we achieve a comparable result in scene 97, and get better results in other scenes except scene 83. The reason is that there may be wrong parameter settings in the code, which makes us not manage to reproduce a 1.01 or similar result in scene 83 using NeuS. 
As for Geo-Neus, we can not reproduce the results reported in the original papers, hence, we train it and ours using the same data for fair comparison. Our results with Geo-Neus are the best among the results of all other state-of-the-art methods. We further provide visual comparisons in Fig.~\ref{fig:error_maps}. Our advantages lie in the smooth surfaces with geometry details. Our methods can leverage more multi-view consistency to infer the implicit functions on the surface.

\begin{table*}[bt]
\centering
\resizebox{\linewidth}{!}{
\begin{tabular}{c | c c c c c c c c c c c c c c c | c}
\Xhline{1.5pt}
Scan & 24 & 37 & 40 & 55 & 63 & 65 & 69 & 83 & 97 & 105 & 106 & 110 & 114 &118 & 122 & Mean \\ 
\hline
NeRF~\cite{mildenhall2020nerf} & 1.90 & 1.60 & 1.85 & 0.58 & 2.28 & 1.27 & 1.47 & 1.67 & 2.05 & 1.07 & 0.88 & 2.53 & 1.06 & 1.15 & 0.96 & 1.49 \\
VolSDF~\cite{yariv2021volume} & 1.14 & 1.26 & 0.81 & 0.49 & 1.25 & 0.70 & 0.72 & 1.29 & 1.18 & 0.70 & 0.66 & 1.08 & 0.42 & 0.61 & 0.55 & 0.86 \\
HF-NeuS~\cite{yiqunhfSDF} & 0.76 & 1.32 & 0.70 & 0.39 & 1.06 & 0.63 & 0.63 & 1.15 & 1.12 & 0.80 & 0.52 & 1.22 & 0.33 & 0.49 & 0.50 & 0.77 \\
MonoSDF~\cite{Yu2022MonoSDF}&0.66&0.88&0.43&0.40&0.87&0.78&0.81&1.23&1.18&0.66&0.66&0.96&0.41&0.57&0.51&0.73\\
\hline
UNISURF~\cite{Oechsle2021ICCV} & 1.32 & 1.36 & 1.72 & 0.44 & 1.35 & 0.79 & 0.8 & 1.49 & 1.37 & 0.89 & 0.59 & 1.47 & 0.46 & 0.59 & 0.62 & 1.02 \\
Ours(UNISURF) & \textbf{0.85} & \textbf{0.95} & \textbf{1.00} & \textbf{0.38} & \textbf{1.25} & \textbf{0.59} & \textbf{0.69} & \textbf{1.36} & \textbf{1.19} & \textbf{0.71} & \textbf{0.52} & \textbf{1.15} & \textbf{0.42} & \textbf{0.48} & \textbf{0.50} & \textbf{0.80} \\
\hline
NeuS~\cite{neuslingjie} & 1.37 & 1.21 & 0.73 & 0.40 & 1.20 & 0.70 & 0.72 & \textbf{1.01} & \textbf{1.16} & 0.82 & 0.66 & 1.69 & 0.39 & 0.49 & 0.51 & 0.87 \\
Ours(NeuS) & \textbf{0.71} & \textbf{0.90} & \textbf{0.68} & \textbf{0.38} & \textbf{1.0} & \textbf{0.60} & \textbf{0.58} & 1.40 & 1.17 & \textbf{0.78} & \textbf{0.52} & \textbf{1.07} & \textbf{0.32} & \textbf{0.43} & \textbf{0.45} & \textbf{0.73} \\
\hline
NeuralWarp~\cite{DBLP:conf/cvpr/DarmonBDMA22} & \textbf{0.49} & 0.71 & 0.38 & 0.38 & 0.79 & 0.81 & 0.82 & 1.20 & \textbf{1.06} & 0.68 & 0.66 & 0.74 & 0.41 & 0.63 & 0.51 & 0.68 \\
Ours(NeuralWarp) & \textbf{0.49} & \textbf{0.68} & \textbf{0.37} & \textbf{0.36} & \textbf{0.73} & \textbf{0.76} & \textbf{0.77} & \textbf{1.17} & 1.10 & \textbf{0.67} & \textbf{0.62} & \textbf{0.65} & \textbf{0.36} & \textbf{0.57} & \textbf{0.49} & \textbf{0.65} \\

\hline
Geo-NeuS~\cite{GEOnEUS2022} & 0.46 & 0.85 & 0.38 & 0.43 & \textbf{0.89} & \textbf{0.50} & 0.50 & 1.26 & 0.89 & 0.66 & 0.52 & 0.82 & \textbf{0.31} & 0.43 & 0.46 & 0.62 \\

Ours(Geo-NeuS~\cite{GEOnEUS2022}) & \textbf{0.42} & \textbf{0.83} & \textbf{0.38} & \textbf{0.37} & 0.90 & 0.53 & \textbf{0.49} & \textbf{1.25} & \textbf{0.88} & \textbf{0.63} & \textbf{0.50} & \textbf{0.78} & \textbf{0.31} & \textbf{0.41} & \textbf{0.43} & \textbf{0.60} \\

\Xhline{1.5pt}
\end{tabular}
}
\vspace{0.1in}
\caption{Numerical comparisons with the latest methods in DTU dataset.}
\vspace{0.1in}
\label{table:dtu}
\end{table*}

\subsection{Evaluations for Scenes}
\noindent\textbf{Dataset and Metrics. }We evaluate our performance in reconstructing scenes from multi-view images from ScanNet~\cite{dai2017bundlefusion} and Replica~\cite{DBLP:journals/corr/abs-1906-05797}. We follow MonoSDF~\cite{Yu2022MonoSDF} to conduct evaluations using the same cases from these dataset. We also use the same metrics including Chamfer distance, the F-Score with a threshold of 5cm, and normal consistency to measure the error between the reconstructed surface and the ground truth surface.

\noindent\textbf{Baselines. }We use MonoSDF~\cite{Yu2022MonoSDF} as the baseline to evaluate our performance for scenes. It is the latest method for learning neural signed distance functions with depth and normal priors on images.

\noindent\textbf{Details. }MonoSDF normalizes a scene into a cube with a range of $[-3.5,3.5]$, we use a resolution $R=51200$ to produce our results. We use the official code released by MonoSDF to produce our results with discrete coordinates. We use discrete coordinates to calculate positional encodings and also calculate normals at discrete coordinates.

\begin{figure}[tb]
  \centering
   \includegraphics[width=\linewidth]{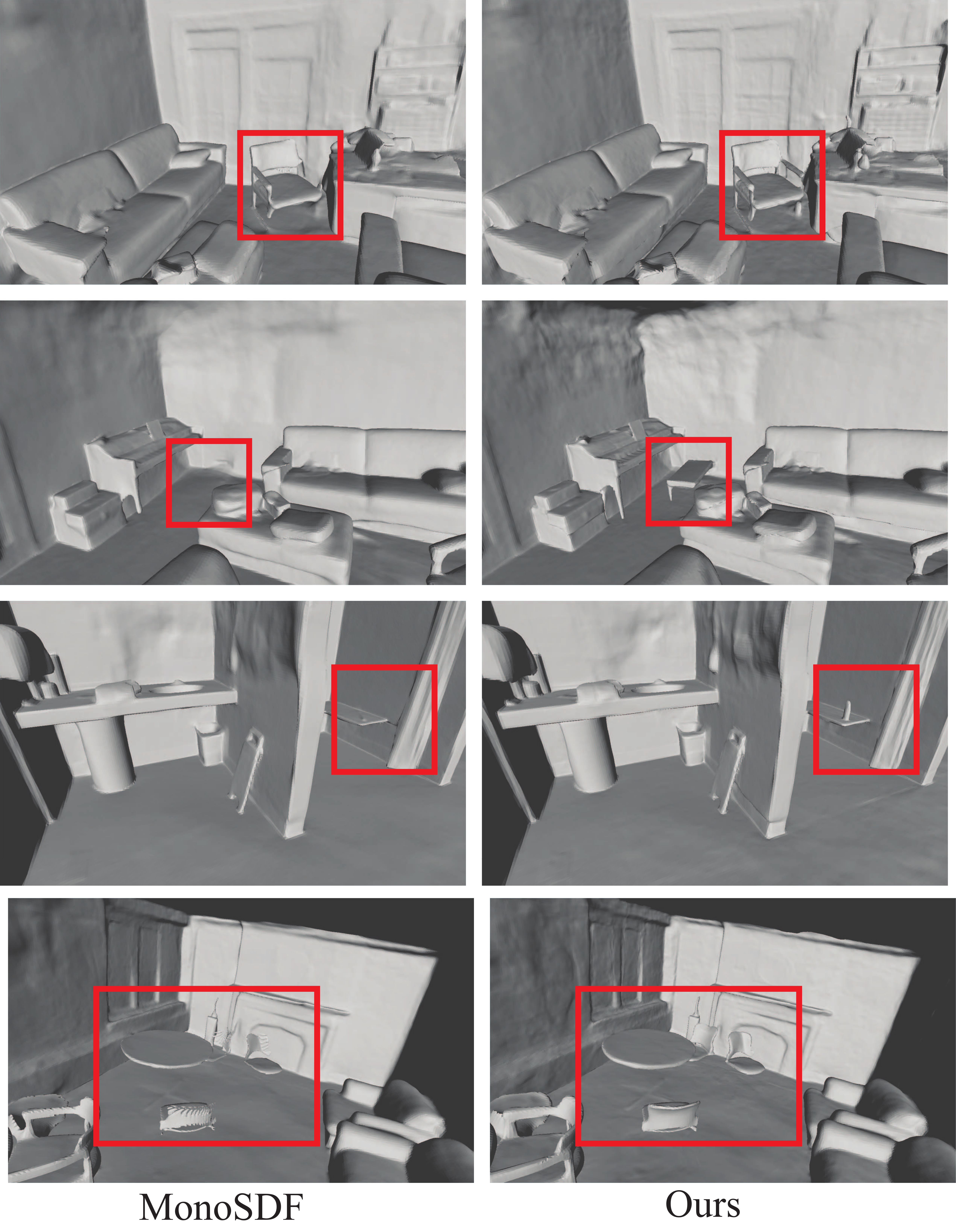}
  %
  %
  \vspace{-0.0in}
\caption{\label{fig:scannet}Visualization of improvements over MonoSDF in ScanNet. More visual comparisons in ScanNet can be found in our supplementary materials.}
\vspace{-0.0in}
\end{figure}

\noindent\textbf{Comparisons. }We report our numerical comparisons in ScanNet in Tab.~\ref{table:ScaneNet} which shows the average results over several scenes. We can see that we achieve much better results than our baseline MonoSDF, especially in terms of the metrics for surface smoothness. We provide the visual comparisons in Fig.~\ref{fig:scannet} where we reconstruct more complete and more accurate surfaces that the other methods.

\begin{table}[h]
\centering
\resizebox{\linewidth}{!}{
\begin{tabular}{c c c c c c c}
\Xhline{1.5pt}
& Chamfer-L1$\downarrow$ & Precision$\uparrow$ & Recall$\uparrow$ & F-score$\uparrow$\\
\hline
COLMAP~\cite{schoenberger2016mvs}  & 0.141 & 0.711 & 0.441 & 0.537\\
UNISURF~\cite{Oechsle2021ICCV}  & 0.359 &0.212& 0.362 &0.267\\
NeuS~\cite{neuslingjie}  & 0.194 & 0.313 & 0.275 & 0.291\\
VolSDF~\cite{yariv2021volume}  & 0.267 & 0.321 & 0.394 & 0.346\\
Manhattan~\cite{guo2022manhattan} & 0.070 & 0.621 & 0.586 & 0.602\\
NeuRIS~\cite{wang2022neuris} & 0.050 & 0.717 & 0.669 & 0.692\\
MonoSDF~\cite{Yu2022MonoSDF} &  0.042&  0.799&  0.681&  0.733\\
\hline
Ours & \textbf{0.039} & \textbf{0.794} & \textbf{0.750} & \textbf{0.770}\\
\Xhline{1.5pt}
\end{tabular}}
\vspace{0.0in}
\caption{Numerical comparisons in ScanNet.}\vspace{0.0in}
\label{table:ScaneNet}
\end{table}

We further report our results in Replica. Our numerical and visual comparisons are shown in Tab.~\ref{table:Replica}. We see that quantized coordinates can significantly reduce the variations in the sample space, and trigger more multi-view consistency at intersections of rays, which leads to more accurate, more completed, and smoother surfaces.

\begin{table}[h]
\centering
\resizebox{0.8\linewidth}{!}{
\begin{tabular}{c c c c}
\hline
& Normal C.$\uparrow$ & CD-L1$\downarrow$ & F-score$\uparrow$\\
\hline
MonoSDF~\cite{Yu2022MonoSDF} & 92.11 & 2.94 & 86.18\\
Ours & \textbf{93.86} & \textbf{2.76} & \textbf{90.16}\\
\hline
\end{tabular}}
\vspace{0.1in}
\caption{Numerical comparison with MonoSDF in Replica.}\vspace{-0.1in}
\label{table:Replica}
\end{table}

\subsection{Analysis}
We provide statistical analysis for our improvements over baselines. With quantized coordinates, we enable to decrease the variations in the sample space and trigger more multi-view consistency by involving the same discrete coordinate in volume rendering along rays from different view. This significantly leaves much less uncertainty and ambiguity in the field, which stabilizes the optimization. We repeat the same procedures as in Fig.~\ref{fig:counts} to count the number of unique coordinates that the network has seen and the number of multi-view consistency that takes effect in the first 1750 iterations using our method and UNISURF. We use the $\log$ function to scale the value, and report the ratio that UNISURF is over us in each scene in DTU in Tab.~\ref{table:countall}. As we can see, UNISURF has to observe 86.2 times more unique coordinates in average than us to infer neural implicit functions, however, can merely use 0.029 the number of multi-view constraints on the intersection of rays from different views of ours. Although neural networks can generalize values at continuous coordinates to the neighboring area, this brings uncertainty and ambiguity in the field, which may cause conflict effect in optimization that results in noisy surfaces and artifacts in empty space.

\begin{table*}[bt]
\centering
\resizebox{\linewidth}{!}{
\begin{tabular}{c | c c c c c c c c c c c c c c c | c}
\Xhline{1.5pt}
Scan & 24 & 37 & 40 & 55 & 63 & 65 & 69 & 83 & 97 & 105 & 106 & 110 & 114 &118 & 122 & Mean \\ [0.5ex]
\hline
Unique Ratio & 140.0 & 50.0 & 163.5 & 80.8 & 57.8 & 99.6 & 48.2 & 59.4 & 107.1 & 89.1 & 86.4 & 43.5 & 136.4 & 75.6 & 55.3 & 86.2 \\
Consistency Ratio
& 0.036 & 0.016 & 0.063 & 0.0234 & 0.012 & 0.024 & 0.021 & 0.014 & 0.027 & 0.024 & 0.053 & 0.022 & 0.041 & 0.039 & 0.022 &0.029\\
\Xhline{1.5pt}
\end{tabular}}
\vspace{0.0in}
\caption{Statistical analysis for our improvements over baselines.}\vspace{-0.1in}
\label{table:countall}
\end{table*}

\subsection{Ablation Studies}
We justify some key modules in our method based on UNISURF in a subset of the DTU dataset. We use Chamfer distance to evaluate the performance.

\begin{figure}[tb]
  \centering
   \includegraphics[width=\linewidth]{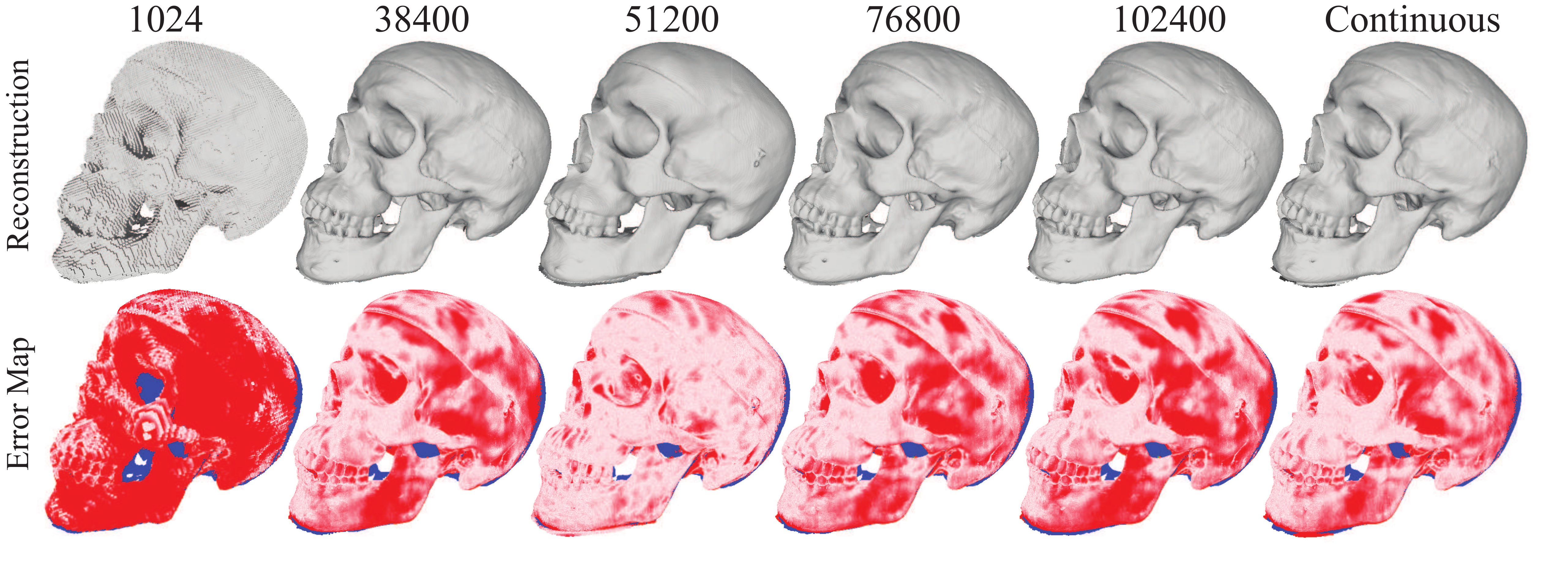}
  %
  %
  \vspace{-0.0in}
\caption{\label{fig:Resolution}Visual comparison of quantized coordinate resolution in terms of reconstruction and error maps.}
\vspace{-0.0in}
\end{figure}

\noindent\textbf{Resolutions. }We explore the effect of resolutions by learning neural implicit representations with quantized coordinates in different resolutions in Tab.~\ref{table:resolution}. We use different resolution candidates $\{1024,25600,38400,51200,76800,102400\}$ to reconstruct the same scene. Compared to the continuous coordinates which can be regarded as infinity high resolution, as shown by ``$\infty$'', we achieve the best performance in $R=51200$. The comparison shows that low resolution does not help infer accurate implicit representations, while the results with a too high resolution approach to the results with continuous coordinates. We visualize the effect of resolution in Fig.~\ref{fig:Resolution}. With low resolution like 1024, we observe severe border inconsistency on the reconstructions and large error in the error map. While the error goes higher if we use a too high resolution. This is because much fewer points sampled along two rays with an intersection can overlap at the same discrete locations. Therefore, a too high resolution may degenerate the result.

\begin{table}[h]
\centering
\resizebox{\linewidth}{!}{
\begin{tabular}{c | c c c c c c c c c c c c c c c | c}
\Xhline{1.5pt}
Resolution &1024&25600 & 38400 & 51200 & 76800 &102400 & $\infty$ \\ 
\hline
Chamfer $\downarrow$ &1.67& 0.70 & 0.64 &\textbf{0.59} & 0.62 & 0.69 & 0.79\\
\Xhline{1.5pt}
\end{tabular}
}
\vspace{0.0in}
\caption{Effect of resolution for quantized coordinates.}
\vspace{-0.00in}
\label{table:resolution}
\end{table}

\noindent\textbf{Border Consistency. }Since our quantized coordinates are defined in extremely high resolution, we achieve pretty good consistency on the border of neighboring quantized coordinates even we use the nearest interpolation to discretize continuous coordinates. We justify our nearest interpolation by comparing with the trilinear interpolation. For a query, we use the 8 nearest quantized coordinates to produce 8 occupancy labels and features, which are further used to predict 8 occupancy labels and colors for trilinear interpolation. The result of ``Trilinear'' in Tab.~\ref{table:trilinear} shows that performing trilinear interpolation in extremely high resolution does not make the optimization converge well, and also brings 7 times more computation.

\noindent\textbf{Reconstruction with Marching Cubes. }We explore the effect of discrete coordinates on extracting surfaces with the marching cubes. With a implicit function learned with quantized coordinates, we can use either discrete coordinates or continuous coordinates to reconstruct meshes. The result of ``MarchingCubes'' Table~\ref{table:trilinear} indicates that there is almost no difference between using discrete or continuous coordinates to extract meshes using marching cubes.

\noindent\textbf{Discrete Alternatives. }Besides using discrete coordinates and their corresponding positional encoding at the same time, we explore different discrete alternatives, such as using discrete coordinates with positional encodings of continuous coordinates or using continuous coordinates with positional encodings of discrete coordinates. The comparison in Tab.~\ref{table:trilinear} shows that using discrete coordinates and their corresponding positional encodings achieves the best.

\begin{table}[h]
\centering
\resizebox{\linewidth}{!}{
\begin{tabular}{c | c c c c c c c c c c c c c c c | c}
\Xhline{1.5pt}
Trilinear & Discrete PE & Discrete Coordinates &MarchingCubes& Ours\\
\hline
0.72& 0.75 & 0.74 & 0.586&0.592\\
\Xhline{1.5pt}
\end{tabular}}
\vspace{0.05in}
\caption{Effect of trilinear interpolation and discrete alternatives.}\vspace{-0.1in}
\label{table:trilinear}
\end{table}

\begin{figure}[tb]
  \centering
   \includegraphics[width=\linewidth]{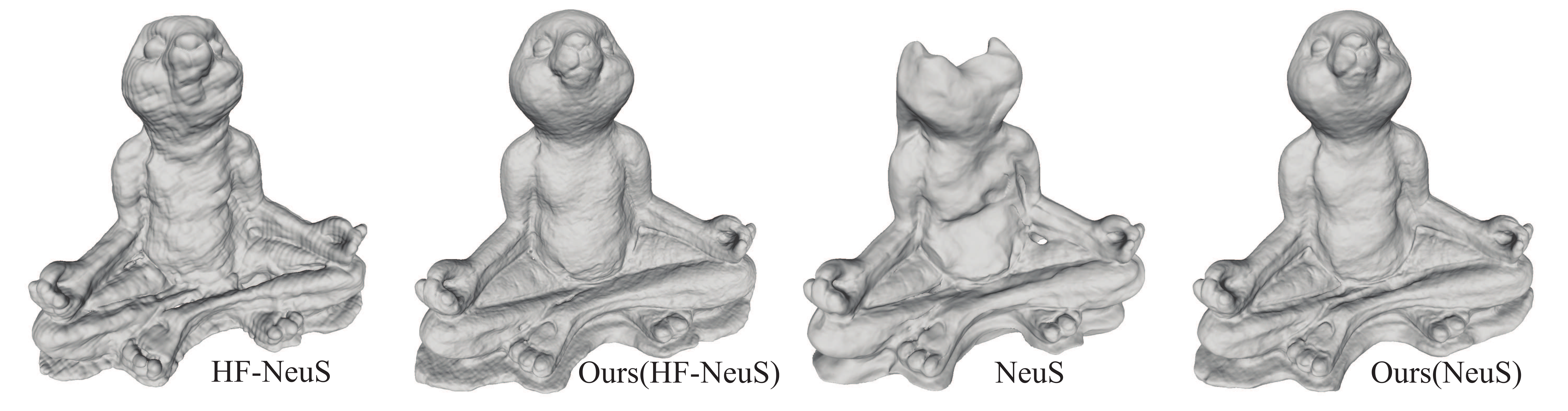}
  %
  %
  \vspace{-0.0in}
\caption{\label{fig:stability}Visualization of Stability with High Frequency.}
\vspace{-0.0in}
\end{figure}

\noindent\textbf{Stability with Higher Frequency.} One advantage of quantized coordinates is to stabilize the optimization with high frequency positional encoding. We conduct experiments to compare with SAPE~\cite{hertz2021sape} using HF-NeuS,  The comparisons in Tab.~\ref{table:pefre} show that our method, working along with HF-NeuS, can further stabilize with positional encoding with higher frequency. In contrast,  HF-NeuS as well as NeuS by themselves are sensitive to the frequency and drastically degenerates its performance on some objects of DTU, as shown in Fig.~\ref{fig:stability}. We found our method can also outperform SAPE in stabilizing optimization with higher frequency. 

\begin{figure}[tb]
  \centering
   \includegraphics[width=\linewidth]{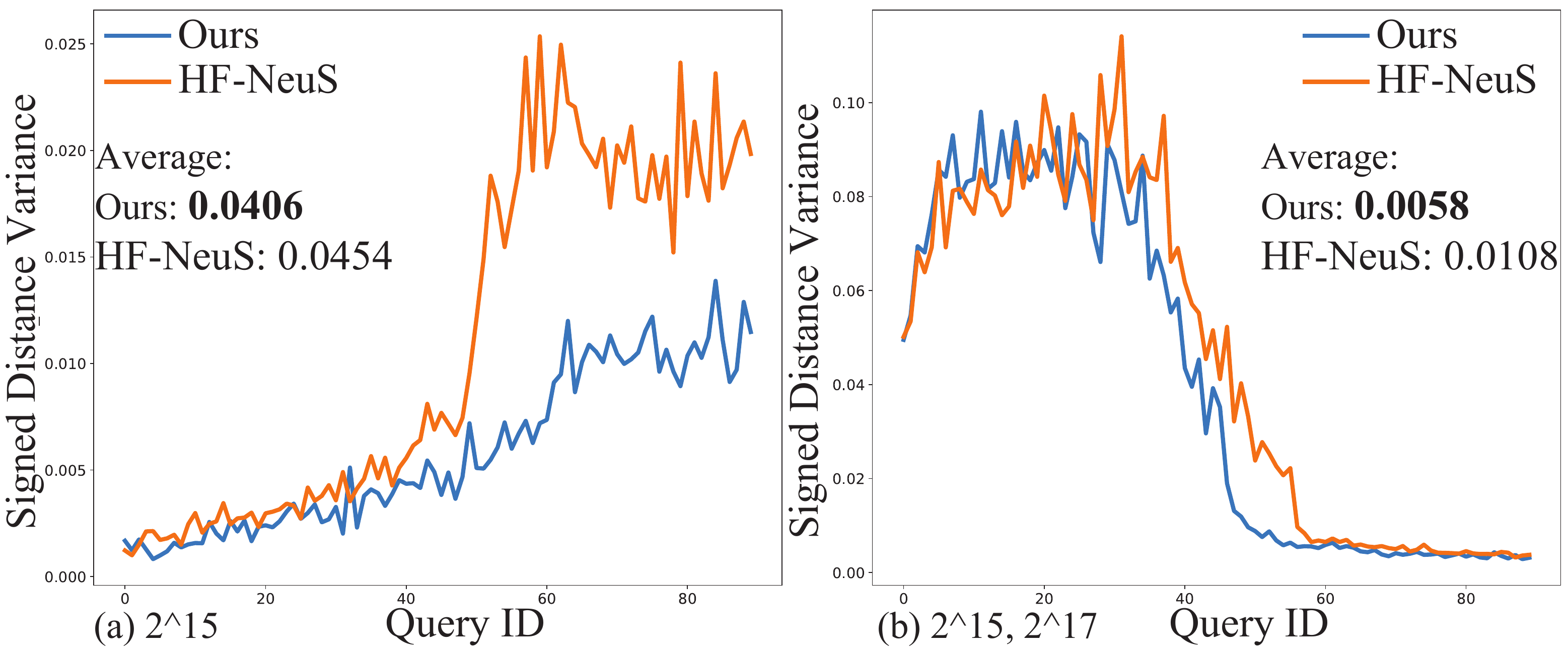}
  %
  %
  \vspace{-0.0in}
\caption{\label{fig:variance}Signed Distance Variance.}
\vspace{-0.0in}
\end{figure}

Moreover, we visualize the signed distance variance change with higher frequencies. We increase the frequency in positional encoding by adding either two high frequency $[2^{14},2^{15}]$ or four high frequencies $[2^{14},...,2^{17}]$. With an interval of $50$ iterations, we record the signed distances predicted by HF-NeuS and ours at the fixed $100$ locations that are randomly sampled on the GT surface. Fig.~\ref{fig:variance} shows the variance of singed distances over $5000$ iterations at each location. The comparisons show that our method produces lower signed distance variance than HF-NeuS on the sampled locations, which indicates that our method stabilizes the learning of signed distances with extremely high frequency positional encoding during training.

\begin{figure}[tb]
  \centering
   \includegraphics[width=\linewidth]{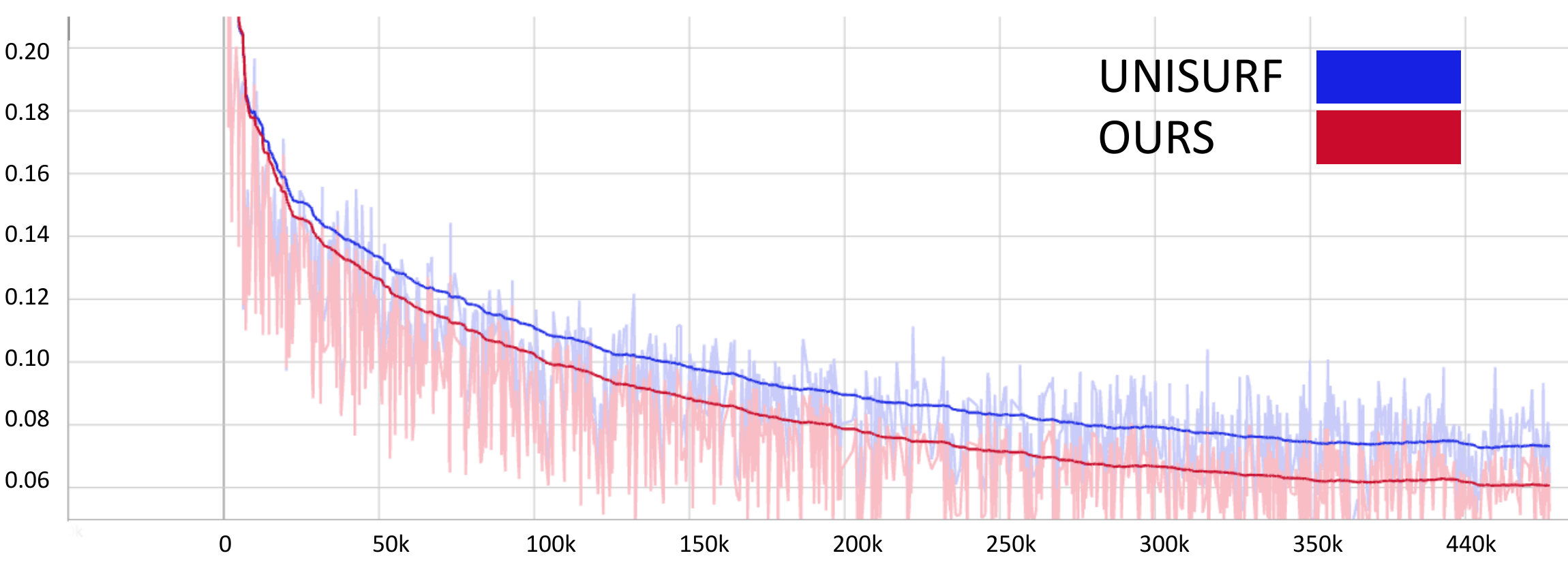}
  %
  %
  \vspace{-0.0in}
\caption{\label{fig:convergence}Speed up the convergence.}
\vspace{-0.0in}
\end{figure}

\noindent\textbf{Runtime Comparisons. }Runtime comparisons in Tab.~\ref{table:runtime} show that our quantized coordinates almost do not bring extra time cost. While quantized coordinates indeed lower the loss, which makes the optimization converge faster, as the comparison with UNISURF in Fig.~\ref{fig:convergence}.


\begin{table}[h]
\centering
\resizebox{\linewidth}{!}{
\begin{tabular}{c | c c c c c}
\Xhline{1.5pt}
Scan & 69 & 83 & 97 & 110 & Mean\\ 
\hline
NeuS~\cite{neuslingjie} &\textbf{0.57} & 1.48&\textbf{1.09} &1.2& 1.09\\
\hline
HF-NeuS~\cite{yiqunhfSDF} & 0.70&1.41 &1.29 &1.58& 1.25\\
\hline
HF-NeuS~\cite{yiqunhfSDF} + OURS & 0.59& \textbf{1.35}& 1.13& \textbf{1.12} & \textbf{1.05}\\
\Xhline{1.5pt}
\end{tabular}}
\vspace{0.1in}
\caption{Effect of stabilizing optimization with high frequencies.}\vspace{0.1in}
\label{table:pefre}
\end{table}


\begin{table}[tb]
\vspace{-0.00in}
\centering
\resizebox{\linewidth}{!}{
\begin{tabular}{c |c| c |c |c| c}
${}$& UNISURF & NeuS & HF-NeuS & NeuralWarp & Geo-NeuS \\
\hline
baseline & 656.221 & 104.726 &  339.816 & 114.312 & 159.237 \\
\hline
Ours & 656.992 & 104.591 & 340.214 & 114.411 & 158.819\\

\end{tabular}}
\vspace{-0.00in}
\caption{Comparison of 1000 iters run-time over baseline.}
\label{table:runtime}
\vspace{-0.00in}
\end{table}

\section{Conclusion}
We introduce to learn neural implicit functions with quantized coordinates to decrease the uncertainty and ambiguity in the field during the optimization for multi-view 3D reconstruction. We transform continuous coordinates into discrete ones using nearest interpolation over the quantized coordinates. Our method significantly stabilizes the optimization and reveal more geometry details with high frequency positional encodings. We successively achieve this by reducing the variations in the sample space and triggering more multi-view consistency constraints to take effect in a more effective way. Our quantized coordinates are defined in extremely high resolution, which however does not bring any extra computational burden or inconsistency on borders of neighboring coordinates. Our experimental results show that we achieve the-state-of-the-art, and justify our ability of improving the accuracy of neural implicit functions learned by different methods in a general way.

\section{Acknowledgements} 
This work was supported by NSF 61972353, IIS-1816511, OAC-1910469, AND OAC-2311245, and Richard Barber interdisciplinary Research Award.
{\small
\bibliographystyle{ieee_fullname}
\bibliography{papers}
}

\end{document}